%% file: ltexpprt_doublecolumn.tex
\documentclass[twoside,leqno,twocolumn]{article}

\usepackage[letterpaper]{geometry}

\usepackage{ltexpprt}
\usepackage{hyperref}

\usepackage[numbers]{natbib}
\usepackage{algorithm}
\usepackage{algorithmic}
\usepackage{amsmath}
\usepackage{amssymb}
\usepackage{adjustbox}
\usepackage{caption}
\usepackage{subcaption}
\usepackage{xcolor}
\usepackage{amsmath}
\usepackage{amsfonts}
\usepackage{bm}
\usepackage{multirow}
\usepackage{makecell}
\usepackage{booktabs}
\usepackage{dsfont}
\usepackage{tablefootnote}
\usepackage{mathrsfs}
\usepackage{pifont}
\usepackage{bbm}
\newcommand{\xmark}{\ding{55}}

\begin{document}

\newcommand\relatedversion{}
\renewcommand\relatedversion{\thanks{The full version of the paper can be accessed at \protect\url{https://arxiv.org/abs/1902.09310}}} % Replace URL with link to full paper or comment out this line

\title{ProtoNAM: Prototypical Neural Additive Models for Interpretable Deep Tabular Learning}

% \author{Anonymous authors}

\author{Guangzhi Xiong\thanks{University of Virginia, \{hhu4zu, ss7mu, aidong\}@virginia.edu}
\and Sanchit Sinha$^*$
\and Aidong Zhang$^*$}

\date{}

\maketitle

% Copyright Statement
% When submitting your final paper to a SIAM proceedings, it is requested that you include
% the appropriate copyright in the footer of the paper.  The copyright added should be
% consistent with the copyright selected on the copyright form submitted with the paper.
% Please note that "20XX" should be changed to the year of the meeting.

% Default Copyright Statement
% \fancyfoot[R]{\scriptsize{Copyright \textcopyright\ 2025 by SIAM\\
% Unauthorized reproduction of this article is prohibited}}

% Depending on which copyright you agree to when you sign the copyright form, the copyright
% can be changed to one of the following after commenting out the default copyright statement
% above.

%\fancyfoot[R]{\scriptsize{Copyright \textcopyright\ 20XX\\
%Copyright for this paper is retained by authors}}

%\fancyfoot[R]{\scriptsize{Copyright \textcopyright\ 20XX\\
%Copyright retained by principal author's organization}}

%\pagenumbering{arabic}
%\setcounter{page}{1}%Leave this line commented out.

\input{content/abstract}

\input{content/introduction}
\input{content/background}
\input{content/methodology}
\input{content/experiment}

\input{content/conclusion}

\bibliographystyle{plain}
\bibliography{reference}

\appendix
\newpage
\input{content/appendix}

\end{document}

%% file: content/abstract.tex
\begin{abstract}
Generalized additive models (GAMs) have long been a powerful white-box tool for the intelligible analysis of tabular data, revealing the influence of each feature on the model predictions. Despite the success of neural networks (NNs) in various domains, their application as NN-based GAMs in tabular data analysis remains suboptimal compared to tree-based ones, and the opacity of encoders in NN-GAMs also prevents users from understanding how networks learn the functions. In this work, we propose a new deep tabular learning method, termed Prototypical Neural Additive Model (ProtoNAM), which introduces prototypes into neural networks in the framework of GAMs. With the introduced prototype-based feature activation, ProtoNAM can flexibly model the irregular mapping from tabular features to the outputs while maintaining the explainability of the final prediction. We also propose a gradient-boosting inspired hierarchical shape function modeling method, facilitating the discovery of complex feature patterns and bringing transparency into the learning process of each network layer. Our empirical evaluations demonstrate that ProtoNAM outperforms all existing NN-based GAMs, while providing additional insights into the shape function learned for each feature. The source code of ProtoNAM is available at \url{https://github.com/Teddy-XiongGZ/ProtoNAM}.

\end{abstract}

%% file: content/introduction.tex
\section{Introduction}

In the field of machine learning, generalized additive models (GAMs) have been recognized for their interpretability \cite{chang2021interpretable,hastie1995generalized,hegselmann2020evaluation,izadi2021generalized,pedersen2019hierarchical,sapra2013generalized}. It provides a transparent perspective through which the impact of individual features on predictions can be observed. This transparency is achieved by independently modeling each feature with a non-linear transformation, thus providing an intuitive understanding of the data. With the evolution of machine learning, sophisticated modeling tools such as Gradient Boosting Decision Trees (GBDTs) and neural networks (NNs) have been integrated with GAMs. This integration aims to harness the expressive power of these advanced models while preserving the interpretability that is the hallmark of GAMs, seeking to strike a balance between expressiveness and interpretability.

Despite the great success of deep neural networks (DNNs) in areas such as computer vision \cite{kang2023neural,khan2018guide,krizhevsky2017imagenet,zhao2024review} and natural language processing \cite{devlin2018bert,khurana2023natural,min2023recent,young2018recent}, they fail to establish a definitive advantage over GBDTs in the area of tabular data analysis \citep{grinsztajn2022why,mcelfresh2024neural}. Tabular data is prevalent in high-stakes domains such as finance and healthcare, where numerical features may exhibit complex patterns due to their potential correlation with discrete categorical features (e.g., age and retirement status). Capturing these patterns is particularly challenging in the context of GAM, where information is encoded independently for each tabular feature. Neural networks, which tend to smooth learned functions, may struggle in modeling these intricate relationships, resulting in their performance often being inferior to GBDT-based methods \citep{grinsztajn2022why}.

However, while GBDT-based GAMs maintain their lead in performance, they are not as flexible as NNs in terms of multitask modeling. Neural networks exhibit unique capabilities in multi-task learning \citep{agarwal2021neural,ruder2017overview}, fitting multiple tasks simultaneously with one network, which can help mitigate model bias and thus improve AI fairness. In practice, it is crucial to handle sensitive features such as gender and race with care, as they may not be suitable to serve as direct predictors. Instead, it would be appropriate to treat the prediction for each group as a separate task and design the training on the whole dataset as multitask learning, which can hardly be performed by GBDT-GAMs. 

In response to these challenges, we introduce ProtoNAM, an interpretable deep tabular learning architecture that improves the performance of NN-based GAMs on tabular data using prototypes. ProtoNAM incorporates a novel prototype-based feature activation, which captures representative feature values from data and learns the corresponding activation patterns, facilitating the learning of subtle variations of model predictions in the feature domain. We also propose a hierarchical shape function modeling method inspired by gradient boosting, which further reveals the intrinsic mechanism of shape function learning by providing layer-wise explanations. This combination enables ProtoNAM to effectively capture the nuanced patterns present in tabular data while maintaining a high degree of interpretability.

%% file: content/background.tex
\section{Related Work}
\subsection{Machine Learning for Tabular Data.}
In the realm of machine learning, the handling of tabular data has traditionally been dominated by tree-based methods, such as Gradient Boosted Decision Trees (GBDTs) \citep{friedman2001greedy} and their highly optimized implementations like XGBoost \citep{chen2016xgboost}. These methods have proven effective due to their ability to capture non-linear interactions and complex patterns inherent in tabular datasets. Despite the attempt to use neural network (NN)-based methods for tabular data that incorporate mechanisms such as contrastive learning \citep{bahri2021scarf} and attention-based pre-training \citep{somepalli2021saint}, neural networks have not yet consistently surpassed the performance of their tree-based counterparts on tabular data \citep{bahri2021scarf,chen2022danets,shwartz2022tabular}. This shortfall is attributed to several factors, as outlined by \cite{grinsztajn2022why}, among which neural networks' inherent bias towards learning overly smooth functions is a critical one. Such an issue is particularly pronounced in the context of GAMs, where each feature is encoded to pursue a nuanced understanding of the underlying knowledge without information from any other factors. Consequently, the challenge remains to develop neural network architectures that can effectively model the complex and often discrete patterns present in tabular data.

\subsection{Generalized Additive Models.}

Generalized Additive Models (GAMs) \citep{hastie2017generalized} are a class of interpretable models where the impact of individual features on the final output can be independently modeled and visualized. These models, which explicitly present the influence of individual features in an additive manner, have been extended and enhanced through various innovative approaches \cite{agarwal2021neural,chang2022node,lou2012intelligible,radenovic2022neural,zhang2024gaussian}. The Explainable Boosting Machine (EBM) \citep{lou2012intelligible} exemplifies this evolution by incorporating boosting techniques to refine the accuracy of traditional GAMs without compromising their interpretability. Based on it, the Neural Oblivious Decision Ensembles for GAM (NODE-GAM) \citep{chang2022node} integrates the principles of GAMs with the NODE architecture \citep{popov2020neural}, leveraging differentiable decision trees to enhance interpretability while maintaining the model's capacity to capture complex patterns. The Neural Additive Models (NAMs) \citep{agarwal2021neural} further advance this field by employing neural networks in the encoding of each feature, which starts the exploration of NN-based GAMs. Building upon the foundation laid by NAMs, Neural Basis Models (NBMs) \citep{radenovic2022neural} have been proposed to reduce the complexity of these models by learning shared basis functions, thus striking a balance between parameter efficiency and interpretability. While \cite{agarwal2021neural} demonstrated the unique capabilities of NN-based GAMs on multitask learning, the performance of existing NN-based GAMs is not on par with their tree-based counterparts, which results in an imprecise interpretation of feature attributions and limits their use in real-world scenarios.

\subsection{Prototype-based Neural Network.}
Prototype-based approaches provide case-based explanations that explain a model with sample representatives learned from the dataset. There are some existing works that tried to incorporate prototypes in machine learning models for interpretable classification \cite{bien2011prototype,kim2014bayesian}.  \cite{snell2017prototypical} and \cite{li2018deep} proposed to explain each data point as a combination of learnable prototypes. Specifically, \cite{snell2017prototypical} consider prototypical representation of a class as a means for classification, essentialy learning a metric space partitioned by prototypes. Li et al. \cite{li2018deep} learn prototypes in the latent space, explaining each sample by looking at the model's learned weights on different prototypes. 
However, the use of prototypes in GAMs for tabular data is still under-explored.

%% file: content/methodology.tex
\section{Methodology}

\subsection{Problem Formulation.} \label{sec:structure}

\begin{figure*}[h!]
    \centering
    \includegraphics[width=0.9\linewidth]{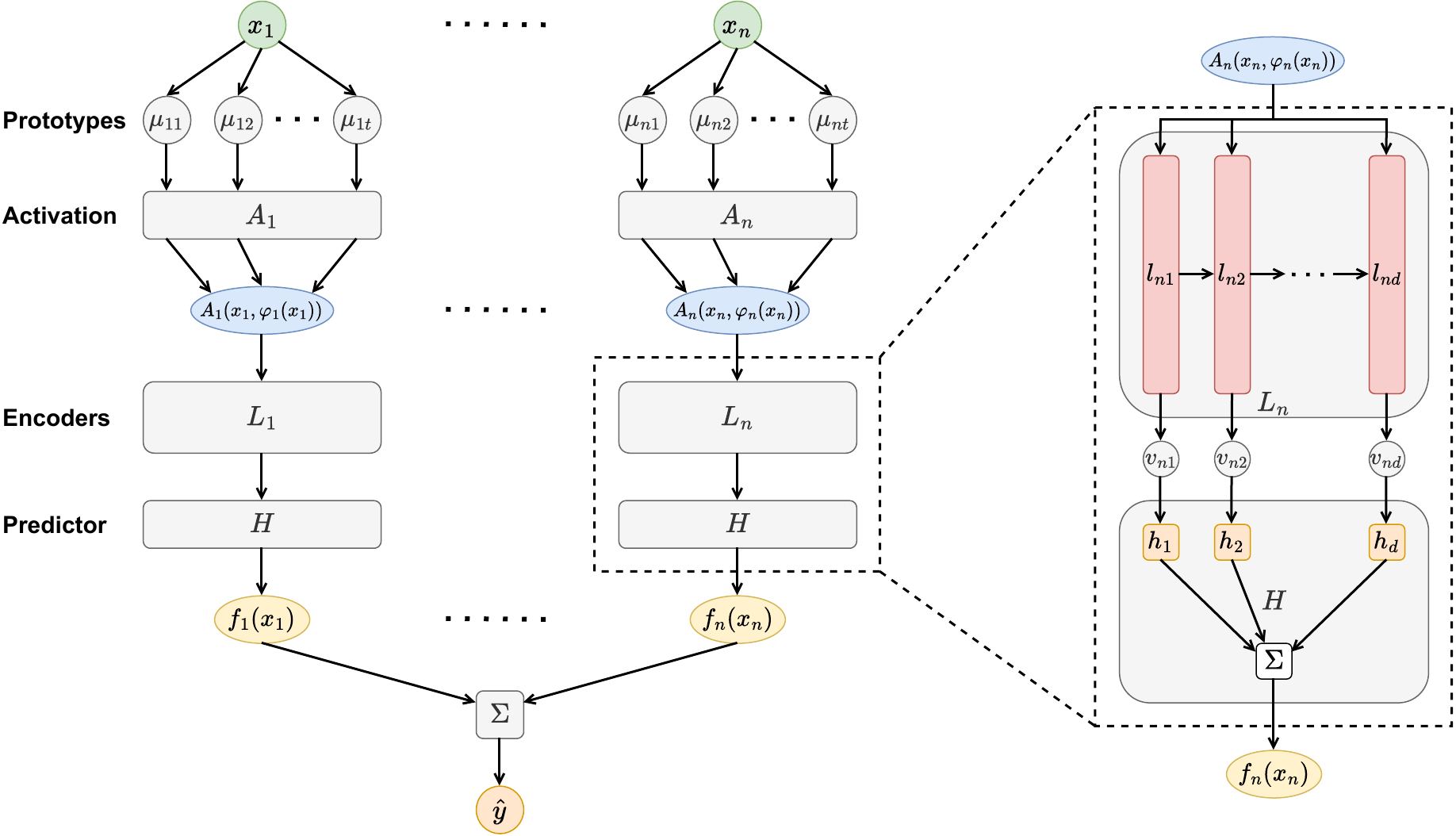}
    \caption{Architecture of Prototypical Neural Additive Model (ProtoNAM). Circles/ovals represent scalar values and rounded rectangles represent functions.}
    \label{fig:architecture}
\end{figure*}

Given a tabular sample with \(n\) features \(x_1,\cdots,x_n\), the task of a GAM, in general, is to learn corresponding feature encoders \(f_1, \cdots, f_n\) which map the input values from feature domains to the prediction domain. The final predicted value of the target output \(y\) is computed as
\begin{equation} \label{eq:y_hat}
    \hat{y} = w_0 + \sum_{i=1}^n w_i f_i(x_i),
\end{equation}
where \(w_1, \cdots, w_n\) are the weights for different features, indicating their relevance in the final prediction. An additional bias term \(w_0\) is also introduced as a learnable inductive bias of the output distribution.

To better capture localized information in tabular features as tree-based GAMs do, we consider a specific task for our NN-based GAM where each encoder \(f_i\) takes both the original input \(x_i\) and the local information \(\varphi_i(x_i)\). \(\varphi_i\) maps \(x_i\) to the localized information by indicating which range the value is on in the domain of the \(i\)-th feature. The process can be formulated as
\begin{equation}\label{eq:g}
    f_i(x_i) = g_i(x_i, \varphi_i(x_i)).
\end{equation}

A unique capability of NN-based GAMs is to perform multitask learning. Suppose the tabular sample has \(s\) different labels \(y_1,\cdots,y_s\) corresponding to \(s\) downstream tasks on the same set of features. A multitask learner is supposed to estimate
\begin{equation} \label{eq:y_multitask}
\hat{y}_u =  w_{u0} + \sum_{i=1}^n w_{ui} f_i(x_i), \quad u\in\{1,\cdots,s\},
\end{equation}
where \(w_{u0}, w_{u1}, \cdots, w_{un}\) are task-specific bias and weights. Features encoders \(f_1, \cdots, f_n\) are shared across tasks.

It is a common practice to incorporate higher-order feature interactions into consideration for precise prediction with some sacrifice in explainability. For an additive model that considers interactions of at most \(k\), the output is computed as
\begin{equation}\label{eq:y_interaction}
    \hat{y} = w_0 + \sum_{v=1}^{k} \sum_{i_v\in \{1,\cdots,n\}^v}w_{i_v}f_{i_v}(x_{i_v}),
\end{equation}
where \(i_v\) is a set of \(v\) indices corresponding to the interaction of \(v\) features. \(x_{i_v}\) is an abbreviation of the \(v\) features that \(i_v\) represents.

The most commonly used high-order additive model is GA\(^2\)M, which can show the effect of binary feature interactions on the output via heat plots. The interactions of more than two features cannot be visualized on 1-D or 2-D figures and thus are not very interpretable to humans.

\subsection{Prototype-based Feature Activation.} \label{sec:prototype}

The motivation for our prototype-based feature activation is to enable neural networks to capture localized information for precise feature encoding. Specifically, the model will learn $t$ feature prototypes \(\mu_{i1}, \cdots, \mu_{it}\) for the \(i\)-th feature \(x_i\). The value of \(t\) is kept the same for different features for simplicity. The input value \(x_i\) will then be mapped to
\begin{equation}
    \varphi_i(x_i) = \left[\rho\left(\|x_i-\mu_{i1}\|\right), \cdots, \rho\left(\|x_i-\mu_{it}\|\right)\right],
\end{equation}
where \(\forall j\in\{1,\cdots,t\}\),
\begin{equation}\small
    \rho(\|x_i-\mu_{ij}\|) = \exp\left(-\frac{\|x_i-\mu_{ij}\|^2}{2\sigma^2}\right).
\end{equation}
The parameter \(\sigma\) controls the extent of localization, which can change dynamically from 1 to 0 during the training. In our implementation, \(\sigma\) is defined as
\begin{equation}\small \label{eq:delay}
    \sigma = \left(1+\exp\left(\frac{T-T_{max}/2}{\tau}\right)\right)^{-1},
\end{equation}
with \(T\) as the current iteration number and \(T_{max}\) as the maximum iteration number. \(\sigma\) will converge to 0 as  \(T \rightarrow T_{max}\), and a hyperparameter \(\tau\) is introduced to control the convergence speed.

Each component in \(\varphi_i(x_i)\) reflects the distance of the feature value \(x_i\) to one representative feature prototype, and \(\varphi_i(x_i)\) in general shows the location information of \(x_i\) on the feature domain. Based on \(\varphi_i(x_i)\), we propose to provide localized linear activation for the input feature by calculating
\begin{equation}\label{eq:activate} \small
    A_i(x_i, \varphi_i(x_i)) = \frac{\sum_{j=1}^{t}(a_{ij}x_i + b_{ij})\rho(\|x_i-\mu_{ij}\|)}{\sum_{j=1}^t \rho(\|x_i-\mu_{ij}\|)},
\end{equation}
where \(\rho(\|x_i-\mu_{ij}\|) / \sum_{j=1}^t \rho(\|x_i-\mu_{ij}\|)\) is the normalized radial-basis-function (RBF) value for the prototype \(\mu_{ij}\), and \(a_{ij}\) and \(b_{ij}\) are parameters of the local linear activation for values around the prototype. By using the normalized RBF transformation as shown in Formula \ref{eq:activate}, we ensure that every value on the feature domain can be properly activated. When \(\sigma\) is close to 0, the activation of the input value will be determined by the local linear mapping of its closest prototype only.

\subsection{Hierarchical Shape Function Modeling.} \label{sec:boosting}
Though the localized RBF feature activation with prototypes enhances the ability of NN-based GAMs to capture nuanced local information, they still present an inferior performance to GBDT-based ones, as shown in our experiments. Moreover, the DNN structure of each feature encoder is not transparent enough to show how the input value ``jumps'' to the predicted outcome. 

Inspired by the idea of gradient boosting, we propose an innovative hierarchical modeling of shape functions that utilizes each layer in the feature encoder as a weaker learner and encourages the model to capture hierarchical information from the data with different layers. Consider we have a \(d\)-layer neural network as the encoder for the \(i\)-th feature, whose layers are labeled as \(l_{i1}, \cdots, l_{id}\). Given an activated input value \(A_i(x_i, \varphi_i(x_i))\), the encoder can output a sequence of embeddings \(v_{i1}, \cdots, v_{id}\) that are generated by each layer. Formally, the embeddings are computed as
\begin{equation} \label{eq:L}
v_{im} = \left\{
\begin{aligned}
    l_{im}\left(A_i(x_i, \varphi_i(x_i))\right)& \quad \text{ if } m=1,\\
    l_{im}(v_{i(m-1)}, A_i(x_i, \varphi_i(x_i))& \quad \text{ if } m>1,
\end{aligned}
\right.
\end{equation}
where \(l_{i2}, \cdots, l_{id}\) takes both the original activated value and the embedding from the previous layer as the input for in-depth encoding. 

The generated embeddings will then be used to supervise the model for hierarchical shape function modeling. As mentioned above, we treat each layer as a weak learner which is prepared for ``gradient boosting''. To map the embeddings from the latent spaces to the predicted label space, we introduce \(d\) weaker predictors \(h_{i1}, \cdots, h_{id}\), which are NNs with only one or two layers. The training objective is then designed to encourage each layer in the encoder to learn ``pseudo-residuals'' of layers previous to it. Specifically, the \(d\)-layer encoder for the \(i\)-th feature can output \(d\) different encoded values \(f_{i1}',\cdots, f_{id}'\), where the \(m\)-th value is computed as the sum of outputs from the first \(m\) layers:
\begin{equation} \label{eq:H} 
    f_{im}' = h_{i1}(v_{i1}) + \cdots + h_{im}(v_{im}).
\end{equation}
We can then generate \(d\) different predictions based the encoded values:
\begin{equation} \label{eq:y_prime}
    \hat{y}_m' = w_{0m} + \sum_{i=1}^n w_{im}f_{im}', \quad \forall m\in\{1,\cdots,d\}
\end{equation}
where \(w_{0m},w_{1m},\cdots,w_{nm}\) are the bias and weights for the prediction based on the first \(m\) layers. It should be noted that the prediction with \(m=d\) is the exact final prediction that the model generates for the downstream task, i.e., \(f_i(x_i) = f_{id}'\) and \(\hat{y} = \hat{y}_d'\). The function \(g_i\) in Formula \ref{eq:g} can then be rewritten as
\begin{equation} \label{eq:g_update}
    g_i = H_i \circ L_i \circ A_i ,
\end{equation}
where \(L_i\) is the mapping from the activated input \(A_i(x_i, \varphi_i(x_i))\) to \(v_{i1},\cdots,v_{id}\) (Formula \ref{eq:L}) and \(H_i\) maps \(v_{i1},\cdots,v_{id}\) to \(f_i(x_i)\) (Formula \ref{eq:H}). The learned bias and weights \(w_{0d},w_{1d},\cdots,w_{nd}\) in Formula \ref{eq:y_prime} are the corresponding parameters used in Formula \ref{eq:y_hat}. 

To encourage the learning of hierarchical information, the training objective is set as
\begin{equation}
    \text{minimize} \quad \sum_{m=1}^d\mathcal{L}(y, \hat{y}_m')
\end{equation}
where \(\mathcal{L}\) is the cross-entropy loss for classification tasks and mean square error loss for regression tasks. The objective guides the model to make precise predictions in a hierarchical way by penalizing the prediction errors given by the first \(m\) layers for all \(m\in\{1,\cdots, d\}\). Our empirical results show that each feature encoder in our model does utilize the first few layers to capture the major pattern in the shape function, allowing the subsequent layers to model nuanced variations in the shape plot.

\subsection{Conditional Feature Interaction Modeling.} \label{sec:generalization}

\input{tables/performance}

While the GAMs only learn \(n\) encoders for \(n\) different features, things become complex when it comes to the modeling of feature interactions. As shown in Formula \ref{eq:y_interaction}, for possible interactions of \(k\) feature, a total number of \(n\choose k\) encoders are required, which can be costly when \(n\) or \(k\) is large. For the scalability of high-order feature interaction modeling, we simplify Formula \ref{eq:y_interaction} as follows:
\begin{equation} \label{eq:y_interact_update}
    \hat{y} = w_0 + \sum_{i_k\in \{1,\cdots,n\}^k}w_{i_k}f(x_1,\cdots,x_n; i_k).
\end{equation}
The modeling for interactions of less than \(k\) features is first removed from the model. Then, instead of training different encoders for different interactions, we propose to learn a shared encoding function \(f\) that can generate \textit{conditional} outputs given information of feature indices.

Specifically, features are first activated and encoded independently by \(A_i\) and \(L_i (i\in\{1,\cdots,n\})\) introduced in Formula \ref{eq:g_update}, which maps each feature \(x_i\) to corresponding latent embeddings \(v_{i1},\cdots,v_{id}\). Similar to Formula \ref{eq:H}, the conditional output given by \(f\) can then be formulated as
\begin{equation}
    f(x_1,\cdots,x_n; i_k) = \sum_{m=1}^d h_m(v_{1m},\cdots,v_{nm}; i_k),
\end{equation}
where \(h_m\) is a conditional predictor corresponding to embeddings given by the \(m\)-th layer in all feature encoders. The conditional interaction modeling with \(h_m\) is implemented as
\begin{equation}\small \label{eq:shared_encoder}
\begin{aligned}
    &h_m(v_{1m},\cdots,v_{nm}; i_k)\\
    =&\text{MLP}\left(\text{concat}\left(v_{1m}\cdot \mathbbm{1}(1\in i_k),\cdots,v_{nm}\cdot \mathbbm{1}(n\in i_k)\right)\right).
\end{aligned}
\end{equation}
The indicator function \(\mathbbm{1}(x)\) has the value 1 if \(x\) is true and 0 if \(x\) is false. Formula \ref{eq:shared_encoder} shows the conditional predictor takes a concatenation of \(n\) vectors as the input, which indicates a parameter size of \(O(n)\). However, it can be used to provide conditional estimations for \(n \choose k\) interactions, which makes the architecture of our model scalable for high-order feature interaction modeling.

An overview of our proposed methods is shown in Figure \ref{fig:architecture}, which presents how different components are composed to learn the shape function of a given feature.

%% file: tables/performance.tex
\begin{table*}[h!] \small
    \centering
    \caption{Performance comparison of our proposed ProtoNAM/ProtoNA$^2$M with baselines on the datasets. Results with $\S$ are reported by \cite{radenovic2022neural}. Results with $\dag$ are reported by \cite{chang2022node}.}
    \begin{tabular}{lcccccccccc}
        \toprule
        \bf \multirow{2.5}{*}{Model} & \bf Housing & \bf \makecell{MIMIC-II} & \bf MIMIC-III & \bf Income & \bf \multirow{2.5}{*}{Avg. Rank} \\
        \cmidrule(lr){2-2}
        \cmidrule(lr){3-3}
        \cmidrule(lr){4-4}
        \cmidrule(lr){5-5}
         &  RMSE $\downarrow$ & AUC $\uparrow$ & AUC $\uparrow$ & AUC $\uparrow$ & \\
        \midrule
        Linear & 0.735\scriptsize $\pm$ 0.000$^\S$ & 0.796\scriptsize $\pm$ 0.012 & 0.772\scriptsize $\pm$ 0.009 & 0.900\scriptsize $\pm$ 0.002 & 7.000 \\ 
        {Spline} & 0.568\scriptsize $\pm$ 0.000 & 0.830\scriptsize $\pm$ 0.011 & 0.810\scriptsize $\pm$ 0.006 & 0.917\scriptsize $\pm$ 0.003 & 5.250 \\
        EBM & 0.559\scriptsize $\pm$ 0.000$^{\S}$ & \bf 0.835\scriptsize $\pm$ 0.011$^{\dag}$ & 0.809\scriptsize $\pm$ 0.004$^{\dag}$ & 0.927\scriptsize $\pm$ 0.003$^{\dag}$ & 2.750 \\
        NODE-GAM & 0.558\scriptsize $\pm$ 0.003 & 0.832\scriptsize $\pm$ 0.011$^{\dag}$ & 0.814\scriptsize $\pm$ 0.005$^{\dag}$ & 0.927\scriptsize $\pm$ 0.003$^{\dag}$ & 2.750 \\
        NAM & 0.572\scriptsize $\pm$ 0.005$^\S$ &  0.834\scriptsize $\pm$ 0.013 & 0.813\scriptsize $\pm$ 0.003 & 0.910\scriptsize $\pm$ 0.003 & 4.250 \\
        NBM & 0.564\scriptsize $\pm$ 0.001$^\S$ & 0.833\scriptsize $\pm$ 0.013 & 0.806\scriptsize $\pm$ 0.003 & 0.918\scriptsize $\pm$ 0.003 & 4.500 \\
        ProtoNAM & \bf 0.553\scriptsize $\pm$ 0.001 & 0.834\scriptsize $\pm$ 0.012 & \bf 0.814\scriptsize $\pm$ 0.003 & \bf 0.927\scriptsize $\pm$ 0.003 & \bf 1.500 \\
        \midrule
        EB$^2$M & 0.492\scriptsize $\pm$ 0.000$^\S$ & 0.848\scriptsize $\pm$ 0.012$^{\dag}$ & 0.821\scriptsize $\pm$ 0.004$^{\dag}$ & 0.928\scriptsize $\pm$ 0.003$^{\dag}$ & 2.750 \\
        NODE-GA$^2$M & 0.476\scriptsize $\pm$ 0.007 & 0.846\scriptsize $\pm$ 0.011$^{\dag}$ & 0.822\scriptsize $\pm$ 0.007$^{\dag}$ & 0.923\scriptsize $\pm$ 0.003$^{\dag}$ & 2.750 \\
        NA$^2$M & 0.492\scriptsize $\pm$ 0.008$^\S$ & 0.843\scriptsize $\pm$ 0.012 & \bf 0.825\scriptsize $\pm$ 0.006 & 0.912\scriptsize $\pm$ 0.003 & 4.000 \\
        NB$^2$M & 0.478\scriptsize $\pm$ 0.002$^\S$ & 0.848\scriptsize $\pm$ 0.012 & 0.819\scriptsize $\pm$ 0.010 & 0.917\scriptsize $\pm$ 0.003 & 3.625 \\
        ProtoNA$^2$M & \bf 0.438\scriptsize $\pm$ 0.004 & \bf 0.849\scriptsize $\pm$ 0.012 & 0.820\scriptsize $\pm$ 0.006 & \bf 0.928\scriptsize $\pm$ 0.003 & \bf 1.875 \\
        \bottomrule
    \end{tabular}
    \label{tab:performance}
\end{table*}

%% file: content/experiment.tex
\section{Experiments}

\subsection{Datasets.}
To systematically evaluate our proposed model and compare it with other methods, we choose four tabular datasets that are commonly used in prior research on GAMs \cite{agarwal2021neural,radenovic2022neural,chang2022node}, including Housing \cite{pace1997sparse}, MIMIC-II \cite{saeed2011multiparameter}, MIMIC-III \cite{johnson2016mimic}, and Income \cite{blake1998uci}. 
The statistics and detailed descriptions of the datasets used are presented in Appendix \ref{sec:app_data_desc}.

\subsection{Baseline Models.} \label{sec:exp_baseline}

ProtoNAM is first compared with existing NN-based GAMs. We consider the Neural Additive Models (NAMs) \citep{agarwal2021neural} and Neural Basis Models (NBMs) \cite{radenovic2022neural}, both of which encode features using neural networks but differ in their approach to sharing basis functions. 
We then compare ProtoNAM with tree-based GAMs, including Explainable Boosting Machine (EBM) \citep{lou2012intelligible,nori2019interpretml}, a GBDT-based GAM which utilizes an ensemble of gradient-boosted decision trees for prediction, and NODE-GAM \citep{chang2022node}, which is a GAM adaption of Neural Oblivious Decision Ensembles (NODE) \citep{popov2020neural} that utilizes MLPs to learn the nodes of decision trees for differentiability. In addition to the GAMs compared, we also include the logistic regression and the spline function as two baselines, which can also explain their predictions with feature attributions.

To evaluate the generalizability of our model to high-order feature interaction modeling, we compare ProtoNA\(^2\)M, a generalized ProtoNAM that learns binary feature interactions with the proposed conditional modeling, to the second-order version of the baseline GAMs, including NA$^2$M, NB$^2$M, EB$^2$M, and NODE-GA$^2$M. More details about the implementation details and the evaluation setting of our experiments can be found in Appendix \ref{sec:app_model_implement} and \ref{sec:app_hyper}.

\subsection{Performance of ProtoNAM.}

Our experimental results, as presented in Table \ref{tab:performance}, offer a comprehensive comparison of the proposed models, ProtoNAM and ProtoNA$^2$M, against various baseline methods. 
In Table \ref{tab:performance}, it can be observed that the tree-based GAMs (EBM \& NODE-GAM) indeed outperform existing NN-based ones (NAM \& NBM) on the tabular datasets with significantly lower average ranks. However, with our proposed methods, ProtoNAM, as a new NN-based GAM, demonstrates superior performance compared to state-of-the-art tree-based models. Notably, on the Housing and Income datasets, ProtoNAM achieves the highest performance among all GAMs tested, which is particularly significant given that tree-based GAMs have traditionally shown strong results on these datasets.

ProtoNA$^2$M, as shown in Table \ref{tab:performance}, also exhibits outstanding performance, ranking first on average across the datasets. It outperforms other baselines, including both tree-based ones and NN-based ones, on most datasets. However, an exception is observed with the MIMIC-III dataset, where NN-based models generally achieve better results. Upon closer examination of the MIMIC-III dataset, originally processed and provided by NODE-GAM \citep{chang2022node}, we found that all categorical variables had been transformed into dummy variables, which are well-suited for neural network processing. Although ProtoNA$^2$M is neural network-based, it incorporates shared predictors for conditional modeling to balance the expressiveness and scalability of the model. This design choice may lead to a slight performance decrease, as observed with ProtoNA$^2$M's performance on MIMIC-III, which is marginally lower than that of NA$^2$M. 

Overall, the average rank scores in Table \ref{tab:performance} underscore the expressiveness of ProtoNAM and ProtoNA$^2$M, with ProtoNAM achieving an average rank of 1.500 and ProtoNA$^2$M achieving an average rank of 1.875. These results highlight the effectiveness of our proposed models in capturing complex data relationships while maintaining interpretability, a key advantage in the realm of GAMs for explainable AI.

\subsection{Interpretability of ProtoNAM.} \label{sec:interpret}

While ProtoNAM demonstrates its superior performance over existing GAMs on tabular datasets, we are curious to see if it can capture more nuanced feature patterns in the shape plots than others for more precise feature interpretation. We train an ensemble of 100 ProtoNAMs and visualize the mean-centered predictions on the shape plots as NAM \citep{agarwal2021neural} and NBM \citep{radenovic2022neural} did. The normalized density of feature values is also presented on the plots using different colors of bars in the background.

\begin{figure}[h]
    \centering
    \includegraphics[width=0.7\linewidth]{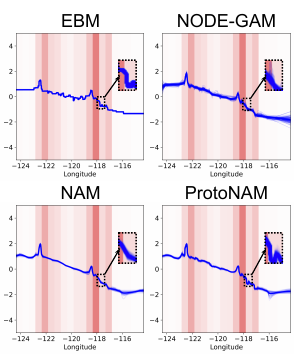}
    \caption{Shape plots of the ``Longitude'' in Housing generated by different GAMs.}
    \label{fig:shape_comp}
\end{figure}

Figure \ref{fig:shape_comp} presents the shape plots for the ``Longitude'' feature in the Housing dataset generated by ProtoNAM and three other baseline GAMs, respectively. Comparing ProtoNAM and NAM, it can be observed that ProtoNAM captures more subtle variations in the shape function than ordinary NN-based models, with a ``jumpy'' curve similar to the ones from tree-based GAMs. The precise modeling of shape plots enables our ProtoNAM to provide a detailed and in-depth explanation of feature attribution. For example, we notice that there is a sudden drop followed by a sharp increase of function values around the Longitude of 117.7$^\circ$W in the shape plot from ProtoNAM. By examining real samples from the dataset, we find that the location corresponds to Laguna Woods Village, which is one of the largest retirement communities in the United States. Since the community is primarily designed for residents aged 55 and older, it could limit the potential buyer pool and thus affect housing prices. However, the pattern is less obvious in EBM and is even hardly found in the plots from NODE-GAM and NAM. Hence, a sharp perception of local variations does help ProtoNAM to present accurate feature attribution interpretations and discover knowledge from real-world data.

\begin{figure}[h]
    \centering
    \includegraphics[width=\linewidth]{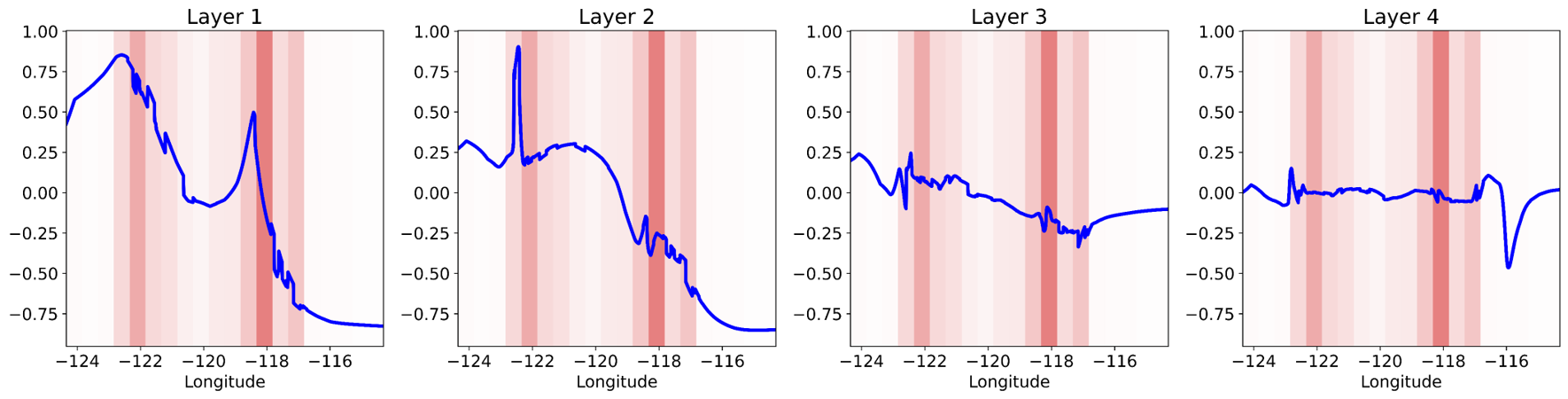}
    \caption{Shape plots of the ``Longitude" feature generated by different layers in ProtoNAM.}
    \label{fig:layers}
\end{figure}

By leveraging a hierarchical structure in the encoder to model shape functions, ProtoNAM can also offer a granular view of the feature influence learned by each layer in the neural network. As depicted in Figure \ref{fig:layers}, the hierarchical decomposition allows us to observe the distinct patterns captured by each layer.
In the initial layer, ProtoNAM adeptly captures the overarching trends and notable inflections, such as the prominent peak at 118.5$^\circ$W. Progressing to the second layer, the model uncovers more nuanced details, like the pronounced jump at approximately 122.5$^\circ$W. Subsequent layers continue to refine the model's understanding, identifying subtle yet critical patterns that contribute to the precision of the predictions. This layer-wise interpretability reveals a diminishing gradient of influence, with the earlier layers playing a more substantial role in shaping the predictions.
By providing insights into the learning process at each layer, ProtoNAM introduces transparency to opaque encoders in NN-based GAMs, enhancing its interpretability of the decision-making process.

\begin{figure}[h]
    \centering
    \includegraphics[width=\linewidth]{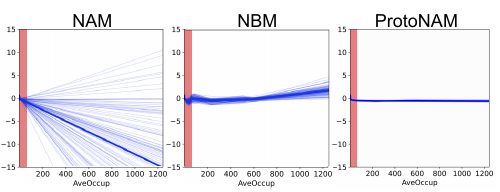}
    \caption{Shape plots of the ``AveOccup'' in Housing generated by different neural network-based GAMs.}
    \label{fig:stability}
\end{figure}

Figure \ref{fig:stability} shows the shape plots of another feature in Housing datasets generated by different NN-based GAMs, including NBM which is described as more stable than its predecessor NAM \cite{radenovic2022neural}. We can see from the shape plots that, among all three NN-based GAMs, ProtoNAM is the most stable one which learns shape functions that are consistent across different trials. 

We attribute this result to the two factors in our ProtoNAM. First, the piece-wise activation function with prototypes has divided the feature domain into pieces, so the model can fit a local shape function at the regions of low data density instead of a global one, which makes it less variable. Second, the proposed hierarchical modeling method encourages the model to learn the main trends of shape function in the first few layers and to use subsequent layers to fit subtle variations, which also prevents the model from learning a variable global function for all the patterns in the shape plots.

\subsection{Multitask Learning of ProtoNAM for Fairness.}

As discussed in Introduction, one of our motivations for the continued exploration of NN-based GAMs is their intrinsic capability as multitask learners, which can be used to address social biases in data from an algorithmic standpoint. To illustrate this, we choose the Income dataset for the case study, where the feature ``Gender'' can be a sensitive input variable. Prior analyses using EBM and NODE-GAM showed that ``Female'' could be a negative predictor of annual income \cite{chang2022node}, raising ethical concerns due to the potential reinforcement or amplification of societal biases and disparities.

To bolster algorithmic fairness with respect to ``Gender'', we implement a multitask learning framework that omits the ``Gender'' feature and instead adjusts the weights of other predictors for different demographic groups as described in Formula \ref{eq:y_multitask}. Specifically, ProtoNAM will learn group-specific weights for its predictors at each layer, corresponding to parameters shown in Formula \ref{eq:y_prime}.

The multitask ProtoNAM achieves a mean AUC score of 0.927 on the test set of the Income dataset with 5-fold cross-validation, which is on par with the one in the original one-task setting. Rather than considering ``Gender'' as a standalone predictor for income estimation, multitask ProtoNAM offers a nuanced approach by providing group-specific feature attributions. This approach not only maintains fairness by excluding sensitive features but also sheds light on the differential impact of other features across demographic groups.

\begin{figure}[h]
    \centering
    \includegraphics[width=0.7\linewidth]{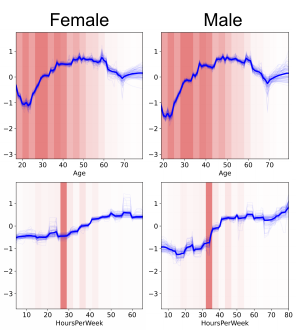}
    \caption{Shape plots of the ``Age'' and ``HoursPerWeek'' features for different groups in the Income dataset generated by ProtoNAM under the multitask learning setting.}
    \label{fig:fairness}
\end{figure}

Figure \ref{fig:fairness} presents the shape plots for two features, ``Age'' and ``HoursPerWeek'', across gender groups generated by the multitask ProtoNAM. The plots reveal that the influence of ``Age'' on income prediction is largely consistent between males and females, with minor variations noted in the early 20s. In contrast, the ``HoursPerWeek'' feature exhibits distinct patterns between the groups. Females are predicted to have higher incomes than males at less than 35 hours per week, but the income prediction for males surges at 35 hours, eventually aligning with that of females. This observation may hint at underlying dynamics within the job market.

The insights provided by ProtoNAM underscore its ability to ensure algorithmic fairness by explicitly excluding sensitive features. It still retains robust performance through the group-wise multitask learning structure, which not only promotes fairness but also offers valuable, group-specific feature attributions that enhance our understanding of the data in a comprehensive and equitable manner.

\subsection{Ablation Studies.} \label{sec:ablation}

In our ablation study, we first investigate the impact of two key aspects of ProtoNAM: the number of prototypes for each feature and the number of layers in the hierarchical shape function modeling. As shown in Table \ref{tab:ablation_layer_proto}, the results reveal that increasing both the number of prototypes and the hierarchical layers leads to improved performance. The increase in the prototype count enables a more nuanced representation of local patterns, which is beneficial for modeling unsmooth relationships within tabular data. Additionally, adding more layers to the hierarchical shape function modeling allows for a more refined capture of intricate data details. Each successive layer builds on the previous, enabling the model to represent increasingly subtle relationships and offering a sophisticated understanding of the data on a deeper level.

\begin{table}[h] \small
    \centering
    \caption{An ablation study of the number of prototypes (\#Prototypes) and the number of layers (\#L) in ProtoNAM on Housing dataset (RMSE $\downarrow$)}
    \begin{tabular}{ccccc}
        \toprule
        \bf \multirow{2.5}{*}{\bf \#L} & \multicolumn{3}{c}{\bf \#Prototypes} \\
        \cmidrule(lr){2-4} 
         & \makecell{8} & \makecell{32} & \makecell{128} \\
        \midrule
        1 & 0.5947\scriptsize $\pm$0.0051 & 0.5881\scriptsize $\pm$0.0048 & 0.5829\scriptsize $\pm$0.0074 \\
        2 & 0.5705\scriptsize $\pm$0.0035 & 0.5669\scriptsize $\pm$0.0032 & 0.5617\scriptsize $\pm$0.0040 \\
        3 & 0.5594\scriptsize $\pm$0.0027 & 0.5584\scriptsize $\pm$0.0033 & 0.5551\scriptsize $\pm$0.0020 \\
        4 & 0.5549\scriptsize $\pm$0.0020 & 0.5533\scriptsize $\pm$0.0013 & \bf 0.5517\scriptsize $\pm$0.0014 \\
        \bottomrule
    \end{tabular}
    \label{tab:ablation_layer_proto}
\end{table}

Moreover, we conduct an ablation study on the individual components of ProtoNAM, namely the prototype-based feature activation and the hierarchical modeling. Results in Table \ref{tab:ablation_components} show that both components are integral to the model, with their absence leading to reduced performance in both ProtoNAM and ProtoNA\(^2\)M. Notably, hierarchical modeling emerges as a more critical factor for performance enhancement in ProtoNAM compared to the prototype-based activation, as the performance drops dramatically with the absence of the former.

\begin{table}[h]
    \centering
    \caption{An ablation study of components in ProtoNAM on Housing dataset (RMSE $\downarrow$). \textcolor{red}{\xmark} denotes the removal of components.}
    \begin{tabular}{lcccc}
        \toprule
        Components & ProtoNAM & ProtoNA$^2$M \\
        \midrule
        full version & \bf 0.553\scriptsize $\pm$0.001 & \bf 0.438\scriptsize $\pm$0.004 \\
        \textcolor{red}{\xmark} hierarchical modeling  & 0.568\scriptsize $\pm$0.007 & 0.440\scriptsize $\pm$0.003 \\
        \textcolor{red}{\xmark} prototypical activation & 0.555\scriptsize $\pm$0.003 & 0.441\scriptsize $\pm$0.005 \\
        \bottomrule
    \end{tabular}
    \label{tab:ablation_components}
\end{table}

\begin{figure}[h]
    \centering
    \includegraphics[width=\linewidth]{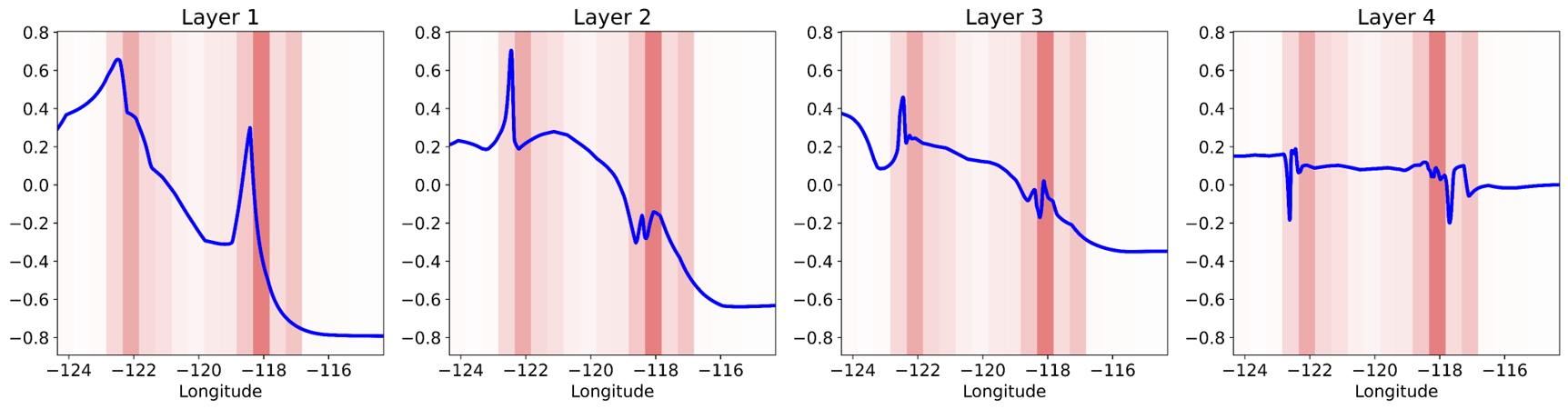}
    \caption{Shape plots of the ``Longitude'' feature generated by ProtoNAM without prototypes.}
    \label{fig:housing_longitude_wo_proto}
\end{figure}

Given the results in Table \ref{tab:ablation_components}, we provide additional visual evidence in Figure \ref{fig:housing_longitude_wo_proto}, which shows the unique importance of the prototype-based feature activation in our model. The layer-wise shape plots for the ``Longitude'' feature in ProtoNAM without prototypes demonstrate a failure to capture the nuanced local variations that are present in the full model, as shown in Figure \ref{fig:layers}. This lack of details can result in a smoother representation and, consequently, a diminished performance.

%% file: content/conclusion.tex
\section{Conclusion}

We propose Prototypical Neural Additive Model (ProtoNAM), a new deep tabular learning method, which introduces prototypes into neural networks in the framework of generalized additive models (GAMs).
Our extensive experiments demonstrate that ProtoNAM not only outperforms existing GAMs in terms of predictive performance on tabular datasets but also provides additional insights into the learned shape functions through its novel prototype-based feature activation and hierarchical shape function modeling. 
With its neural network-based design, ProtoNAM can be an effective multitask learner, improving algorithmic fairness without compromising on performance. 
The ablation studies further validate the importance of each proposed component in enhancing the model's capabilities. 

%% file: content/appendix.tex
\clearpage

\section{Reproducibility}

\subsection{Dataset Description.} \label{sec:app_data_desc}

\ \\
\textbf{Housing.} California Housing (Housing) \citep{pace1997sparse} is a regression task on the observations of median housing prices in all California census blocks, which contains 20,640 samples with 8 different features. We follow NBM \citep{radenovic2022neural} to use sklearn library's API  for data collection.

\textbf{MIMIC-II.} Multiparameter Intelligent Monitoring in Intensive Care (MIMIC)-II \citep{saeed2011multiparameter} is a dataset that can be used for a binary classification task for the prediction of mortality in ICUs. For a fair comparison with previous work, we use the processed data of MIMIC-II from NODE-GAM \citep{chang2022node}, which contains 17 features including 7 categorical ones.

\textbf{MIMIC-III.} MIMIC-III \citep{johnson2016mimic} is a separate version of MIMIC database, which is larger and more comprehensive than MIMIC-II. The dataset that NODE-GAM  \citep{chang2022node} used in their experiments is shared by the authors, in which categorical variables have been converted to dummy variables.

\textbf{Income.} The Adult Income (Income) dataset from the UCI Machine Learning Repository \citep{blake1998uci} defines a binary classification task to predict who has an income of over \$50k a year. We also use the data shared by NODE-GAM \citep{chang2022node} for this task.

Table \ref{tab:app_data_stat} provides some statistical details of the datasets described above. The evaluation metrics for the datasets are chosen following previous work \citep{agarwal2021neural,radenovic2022neural,chang2022node}.

\begin{table*}[h]
    \centering
    \caption{Datasets for model evaluation.}
    \begin{tabular}{llcccc}
        \toprule
        \bf Dataset & \bf Task & \bf samples & \bf features & \bf categorical features & \bf classes\\
        \midrule
        Housing & regression & 20,640 & 8 & 0 & -- \\
        MIMIC-II & classification & 24,508 & 17 & 7 & 2 \\
        MIMIC-III & classification & 27,348 & 57 & 0 & 2 \\
        Income & classification & 32,561 & 14 & 8 & 2 \\
        \bottomrule
    \end{tabular}
    \label{tab:app_data_stat}
\end{table*}

\subsection{Implementation Details.}
\label{sec:app_model_implement}

For a fair comparison with the baselines, we use the released codes from the papers for NODE, NODE-GAM / NODE-GA$^2$M, EBM / EB$^2$M, and NBM / NB$^2$M. As NBM has implemented NAM / NA$^2$M in PyTorch and compared their performance, we adopt their implementation and compare their models with ours. We implemented the Linear model with the sigmoid activation, and MLP models with 4 hidden layers and 64 neural nodes in each layer.

For the implementation of ProtoNAM, we first implement a multi-channel fully connected network which can encode different features with separate neural encoders in parallel.
In addition, the model is designed to learn different prototypes and corresponding activation functions for its input for different layers so that various patterns of prototypes can be captured by the layers. The number of prototypes learned by each layer for each feature is set as the same for simplicity. 

Following NBM \citep{radenovic2022neural}, we split the Housing dataset into training, validation, and test sets, and use min-max normalization for data pre-processing. The models are run 10 times with different seeds on the dataset, and the mean and standard deviation of the results are reported. For the other three datasets, we use the data split files provided by NODE-GAM \citep{chang2022node}, so that the exact same 5-fold cross-validation can be performed. We perform target encoding for categorical features following NODE \citep{popov2020neural} and NODE-GAM \citep{chang2022node} when running the baselines with neural networks, since such features cannot be processed directly by neural networks due to their discontinuity. However, as ProtoNAM can spit the feature domains into different pieces using prototypes, our model can easily handle the categorical features with simple ordinal encoding. We further performed the same quantile transformation on the data as is done by previous works \citep{chang2022node,popov2020neural,radenovic2022neural}.
In the implementation of ProtoNAM, the prototypes are initialized according to the quantiles of the processed data for each feature. All experiments in this paper are run on our A100 GPUs (40 GB / 80 GB).

\subsection{Hyperparameters.} \label{sec:app_hyper}

To be consistent with existing NN-based NAM \citep{agarwal2021neural,radenovic2022neural}, the encoders in ProtoNAM are implemented with MLPs. The hyperparameter \(\tau\) in Formula \ref{eq:delay} is fixed as 16 in our experiments. We random search other hyperparameters in the following ranges:

\begin{itemize}
    \item \# of prototypes: the number of prototypes learned by ProtoNAM for each feature in each layer, sampled from \{1, 2, 4, 8, 16, 32, 64, 128\}.
    \item \# of layers: the number of layers in the encoder, sampled from \{1, 2, 3, 4\}.
    \item \# of layers for predictors: the number of layers in the predictor, sampled from \{1, 2\}.
    \item Hidden dimension: the number of nodes in each layer of either the encoder or the predictor, sampled from \{32, 64\}.
    \item Batch size: the number of samples in each batch, sampled from \{512, 1024, 2048\}.
    \item Max iteration: the number of iterations for training, sampled from \{200, 500, 1000\}.
    \item Learning rate: the speed of gradient descent, sampled from [1e-6, 1e-1].
    \item Weight decay: the coefficient for the L2 normalization on parameters, sampled from [1e-8, 1e-1].
    \item Dropout: the probability of a parameter being replaced as 0 during training, sampled from \{0.0, 0.1, 0.2, 0.3, 0.4, 0.5\}.
    \item Dropout output: the probability of a feature/interaction's output being replaced as 0 during training, sampled from \{0.0, 0.1, 0.2, 0.3\}.
    \item Output penalty: the coefficient for the L2 normalization on the outputs of features / interactions for reducing unimportant ones, sampled from [1e-5, 1e-1].
    \item Normalization: normalization methods used in encoders and predictors, sampled from \{batch\_norm, layer\_norm\}.
\end{itemize}
We decrease the learning rate during training with cosine annealing following NBM \citep{radenovic2022neural}. The best hyperparameters we found for ProtoNAM and ProtoNA$2$M on the datasets are shown in Tables \ref{tab:app_hyperparam_protonam} and \ref{tab:app_hyperparam_protona2m}, respectively.

\begin{table*}[h]
    \centering
    \caption{Hyperparameters for ProtoNAM on all datasets.}
    \begin{tabular}{l|ccccccc}
        \toprule
        Hyperparameter & Housing & MIMIC-II & MIMIC-III & Income \\
        \midrule
        \# of prototypes & 32 & 64 & 1 & 32  \\
        \# of layers & 4 & 4 & 1 & 4 \\
        \# of layers for predictors & 2 & 2 & 1 & 2\\
        Hidden dimension & 64 & 64 & 64 & 64 \\
        Batch size & 2048 & 2048 & 2048 & 2048 \\
        Max iteration & 1000 & 1000 & 200 & 1000 \\
        Learning rate & 2e-4 & 1e-3 & 5e-2 & 1e-3 \\
        Weight decay & 1e-3 & 1e-1 & 1e-8 & 1e-2 \\
        Dropout & 0.0 & 0.4 & 0.0 & 0.0 \\
        Dropout output & 0.0 & 0.1 & 0.0 & 0.0 \\
        Output Penalty & 1e-3 & 1e-2 & 1e-5 & 1e-2 \\
        Normalization & layer\_norm & layer\_norm & batch\_norm & layer\_norm \\
        \bottomrule
    \end{tabular}
    \label{tab:app_hyperparam_protonam}
\end{table*}

\begin{table*}[h]
    \centering
    \caption{Hyperparameters for ProtoNA$^2$M on all datasets.}
    \begin{tabular}{l|ccccccc}
        \toprule
        Hyperparameter & Housing & MIMIC-II & MIMIC-III & Income \\
        \midrule
        \# of prototypes & 32 & 64 & 16 & 32  \\
        \# of layers & 4 & 4 & 4 & 4 \\
        \# of layers for predictors & 2 & 2 & 2 & 2\\
        Hidden dimension & 64 & 64 & 64 & 64 \\
        Batch size & 2048 & 2048 & 512 & 2048 \\
        Max iteration & 1000 & 1000 & 500 & 1000 \\
        Learning rate & 2e-3 & 4e-3 & 1e-2 & 2e-4 \\
        Weight decay & 1e-3 & 5e-3 & 5e-2 & 8e-3 \\
        Dropout & 0.0 & 0.5 & 0.4 & 0.0 \\
        Dropout output & 0.0 & 0.0 & 0.0 & 0.0 \\
        Output Penalty & 1e-2 & 5e-2 & 1e-2 & 1e-3 \\
        Normalization & layer\_norm & layer\_norm & layer\_norm & layer\_norm \\
        \bottomrule
    \end{tabular}
    \label{tab:app_hyperparam_protona2m}
\end{table*}

\section{More Experimental Results} \label{sec:app_more_results}
\subsection{Prototypes learned by ProtoNAM.}

\begin{table*}[ht] \small
    \centering
    \caption{Selected prototypes and activation functions of the ``Longitude'' feature from the Housing dataset learned by ProtoNAM in the first layer.}    \label{tab:case_prototype}
    \small
    \begin{tabular}{ccccccc}
        \toprule
        \multirow{2}{*}{Normalized}& Prototype ($p$) & 0.4004 & 0.5459 & 0.6450 & 0.6461 \\
        \cmidrule(lr){2-6}
        & Activation ($R$) & 1.01$x$ - 4e-3 & 1.01$x$ - 2e-3 & 1.02$x$ + 5e-3 & 1.02$x$ + 9e-3 \\
        \midrule
        \multirow{2}{*}{Original}& Prototype ($p$) & -120.330 & -118.869 & -117.874 & -117.863 \\
        \cmidrule(lr){2-6}
        & Activation ($R$) & 1.01$x$ + 0.877 & 1.01$x$ + 1.061 & 1.02$x$ + 1.958 & 1.02$x$ + 2.486 \\
        \bottomrule
    \end{tabular}
\end{table*}

Table \ref{tab:case_prototype} shows the selected prototypes and their corresponding activation functions learned by ProtoNAM in the first layer for the ``Longitude'' feature from the Housing dataset. Since the features are normalized before being fed into the model, we perform the inverse transformations on the learned parameters to see how they will affect the feature values in the original space. We can see that while the activation functions for -120.330 and -118.869 are similar, the function for -117.874 is different with a significantly larger bias. The difference between the adjacent prototypes of -118.869 and -117.874 exactly corresponds to the sharp drop in the shape plot at around -118.5. Moreover, the difference between activation functions for -117.874 and -117.863 is also reflected by the sharp decrease of the function at the edge of the dark bar in the plot. Our analysis of prototypes reflects that prototypes not only enable the model to learn subtle patterns for precise prediction, but they also provide us a way to explore how various feature values differ in their influence on the prediction.

\subsection{ProtoNA$^n$M -- Full Complexity.}\label{sec:protonanm}

As is introduced in Section \ref{sec:exp_baseline}, we implemented ProtoNA$^n$M for the full complexity of feature interactions, where $n$ corresponds to the number of features for the tabular data. Table \ref{tab:app_full_comp} shows the comparison of ProtoNA$^n$M with all other models. It can be observed from the table that ProtoNA$^n$M is very powerful in modeling complex feature interactions of the data. On MIMIC-II and MIMIC-III datasets where neural network-based models have comparable performance to tree-based models, our model can outperform all other models and achieve the best results. However, on Housing and Income datasets where ProtoNA$^n$M performs slightly worse than tree-based models, it is our ProtoNA$^2$M that performs the best. The results reflect that while ProtoNAM is the most interpretable one in our framework, ProtoNA$^2$M and ProtoNA$^n$M are quite good at modeling complex feature interactions, and the comparison of their results may vary on different data.

\begin{table*}[htb]
    \centering
    \caption{Performance comparison with all baselines on the datasets. Results with $\S$ are reported by \cite{radenovic2022neural}. Results with $\dag$ are reported by \cite{chang2022node}.}
    \begin{tabular}{lcccccccccc}
        \toprule
        \bf \multirow{2.5}{*}{Model} & \bf Housing & \bf \makecell{MIMIC-II} & \bf MIMIC-III & \bf Income & \bf \multirow{2.5}{*}{Avg. Rank} \\
        \cmidrule(lr){2-2}
        \cmidrule(lr){3-3}
        \cmidrule(lr){4-4}
        \cmidrule(lr){5-5}
         &  RMSE $\downarrow$ & AUC $\uparrow$ & AUC $\uparrow$ & AUC $\uparrow$ & \\
        \midrule
        Linear & 0.735\scriptsize $\pm$0.000$^\S$ & 0.796\scriptsize $\pm$0.012 & 0.772\scriptsize $\pm$0.009 & 0.900\scriptsize $\pm$0.002 & 7.000 \\ 
        {Spline} & 0.568\scriptsize $\pm$0.000 & 0.830\scriptsize $\pm$0.011 & 0.810\scriptsize $\pm$0.006 & 0.917\scriptsize $\pm$0.003 & 5.000 \\
        EBM & 0.559\scriptsize $\pm$0.000$^{\S}$ & \bf 0.835\scriptsize $\pm$0.011$^{\dag}$ & 0.809\scriptsize $\pm$0.004$^{\dag}$ & 0.927\scriptsize $\pm$0.003$^{\dag}$ & 2.750 \\
        NODE-GAM & 0.558\scriptsize $\pm$0.003 & 0.832\scriptsize $\pm$0.011$^{\dag}$ & 0.814\scriptsize $\pm$0.005$^{\dag}$ & 0.927\scriptsize $\pm$0.003$^{\dag}$ & 2.625 \\
        NAM & 0.572\scriptsize $\pm$0.005$^\S$ &  0.834\scriptsize $\pm$0.013 & 0.813\scriptsize $\pm$0.003 & 0.910\scriptsize $\pm$0.003 & 4.375 \\
        NBM & 0.564\scriptsize $\pm$0.001$^\S$ & 0.833\scriptsize $\pm$0.013 & 0.806\scriptsize $\pm$0.003 & 0.918\scriptsize $\pm$0.003 & 4.500 \\
        ProtoNAM & \bf 0.553\scriptsize $\pm$0.001 & 0.834\scriptsize $\pm$0.012 & \bf 0.814\scriptsize $\pm$0.003 & \bf 0.927\scriptsize $\pm$0.003 & \bf 1.750 \\
        \midrule
        EB$^2$M & 0.492\scriptsize $\pm$0.000$^\S$ & 0.848\scriptsize $\pm$0.012$^{\dag}$ & 0.821\scriptsize $\pm$0.004$^{\dag}$ & 0.928\scriptsize $\pm$0.003$^{\dag}$ & 2.750 \\
        NODE-GA$^2$M & 0.476\scriptsize $\pm$0.007 & 0.846\scriptsize $\pm$0.011$^{\dag}$ & 0.822\scriptsize $\pm$0.007$^{\dag}$ & 0.923\scriptsize $\pm$0.003$^{\dag}$ & 2.750 \\
        NA$^2$M & 0.492\scriptsize $\pm$0.008$^\S$ & 0.843\scriptsize $\pm$0.012 & \bf 0.825\scriptsize $\pm$0.006 & 0.912\scriptsize $\pm$0.003 & 4.000 \\
        NB$^2$M & 0.478\scriptsize $\pm$0.002$^\S$ & 0.848\scriptsize $\pm$0.012 & 0.819\scriptsize $\pm$0.010 & 0.917\scriptsize $\pm$0.003 & 3.625 \\
        ProtoNA$^2$M & \bf 0.438\scriptsize $\pm$0.004 & \bf 0.849\scriptsize $\pm$0.012 & 0.820\scriptsize $\pm$0.006 & \bf 0.928\scriptsize $\pm$0.003 & \bf 1.875 \\
        \midrule
        XGBoost & \bf 0.443\scriptsize $\pm$0.000$^\S$ & 0.844\scriptsize $\pm$0.012$^{\dag}$ & 0.819\scriptsize $\pm$0.004$^{\dag}$ & \bf 0.928\scriptsize $\pm$0.003$^{\dag}$ & 1.750 \\
        NODE & 0.523\scriptsize $\pm$0.000 & 0.843\scriptsize $\pm$0.011$^{\dag}$ & 0.828\scriptsize $\pm$0.007$^{\dag}$ & 0.919\scriptsize $\pm$0.003$^{\dag}$ & 3.000 \\
        MLP & 0.501\scriptsize $\pm$0.006$^\S$ & 0.833\scriptsize $\pm$0.013 & 0.815\scriptsize $\pm$0.009 & 0.912\scriptsize $\pm$0.004 & 3.750 \\
        ProtoNA$^n$M & 0.451\scriptsize $\pm$0.006 & \bf 0.849\scriptsize $\pm$0.012 & \bf 0.829\scriptsize $\pm$0.006 & 0.923\scriptsize $\pm$0.003 & \bf 1.500 \\
        \bottomrule
    \end{tabular}
    \label{tab:app_full_comp}
\end{table*}

Table \ref{tab:app_ablation_components} shows the ablation studies of components in ProtoNA$^n$M. 
We notice that ProtoNA$^n$M without prototype-based feature activation performs worse than the full version of ProtoNA$^n$M, and the model without the hierarchical modeling performs even better than the full version.
As an uninterpretable model, ProtoNA$^n$M learns quite complex knowledge, since it takes mixed features as input and can model any interaction of features. The best prediction of ProtoNA$^n$M is difficult to represent as a linear combination of predictions given by different layers, which may explain why it performs better without hierarchical modeling.

\begin{table}[h]
    \centering
    \caption{An ablation study of components in ProtoNA$^n$M on Housing dataset (RMSE $\downarrow$). \textcolor{red}{\xmark} denotes the removal of components.}
    \begin{tabular}{lcccc}
        \toprule
        Components & ProtoNA$^n$M \\
        \midrule
        full version & 0.451\scriptsize $\pm$0.006 \\
        \textcolor{red}{\xmark} hierarchical modeling & \bf 0.443\scriptsize $\pm$0.004\\
        \textcolor{red}{\xmark} prototypical activation & 0.456\scriptsize $\pm$0.015 \\
        \bottomrule
    \end{tabular}
    \label{tab:app_ablation_components}
\end{table}

The hyperparameters we used for the experiments of ProtoNA$^n$M are presented in Table \ref{tab:app_hyperparam_protonanm}.

\begin{table*}[h]
    \centering
    \caption{Hyperparameters for ProtoNA$^n$M on all datasets.}
    \begin{tabular}{l|ccccccc}
        \toprule
        Hyperparameter & Housing & MIMIC-II & MIMIC-III & Income \\
        \midrule
        \# of prototypes & 32 & 64 & 64 & 32  \\
        \# of layers & 4 & 4 & 4 & 4 \\
        \# of layers for predictors & 2 & 2 & 2 & 2\\
        Hidden dimension & 64 & 64 & 64 & 64 \\
        Batch size & 2048 & 2048 & 1024 & 2048 \\
        Max iteration & 1000 & 1000 & 500 & 1000 \\
        Learning rate & 2e-4 & 1e-4 & 1e-4 & 1e-5 \\
        Weight decay & 1e-3 & 1e-2 & 1e-3 & 5e-2 \\
        Dropout & 0.0 & 0.5 & 0.5 & 0.0 \\
        Dropout output & 0.0 & 0.0 & 0.0 & 0.0\\
        Output Penalty & 1e-2 & 1e-3 & 1e-4 & 4e-4 \\
        Normalization & layer\_norm & layer\_norm & layer\_norm & layer\_norm \\
        \bottomrule
    \end{tabular}
    \label{tab:app_hyperparam_protonanm}
\end{table*}

\subsection{Effect of Different Numbers of Prototypes.}

\begin{figure*}[h]
    \centering
    \includegraphics[width=0.8\linewidth]{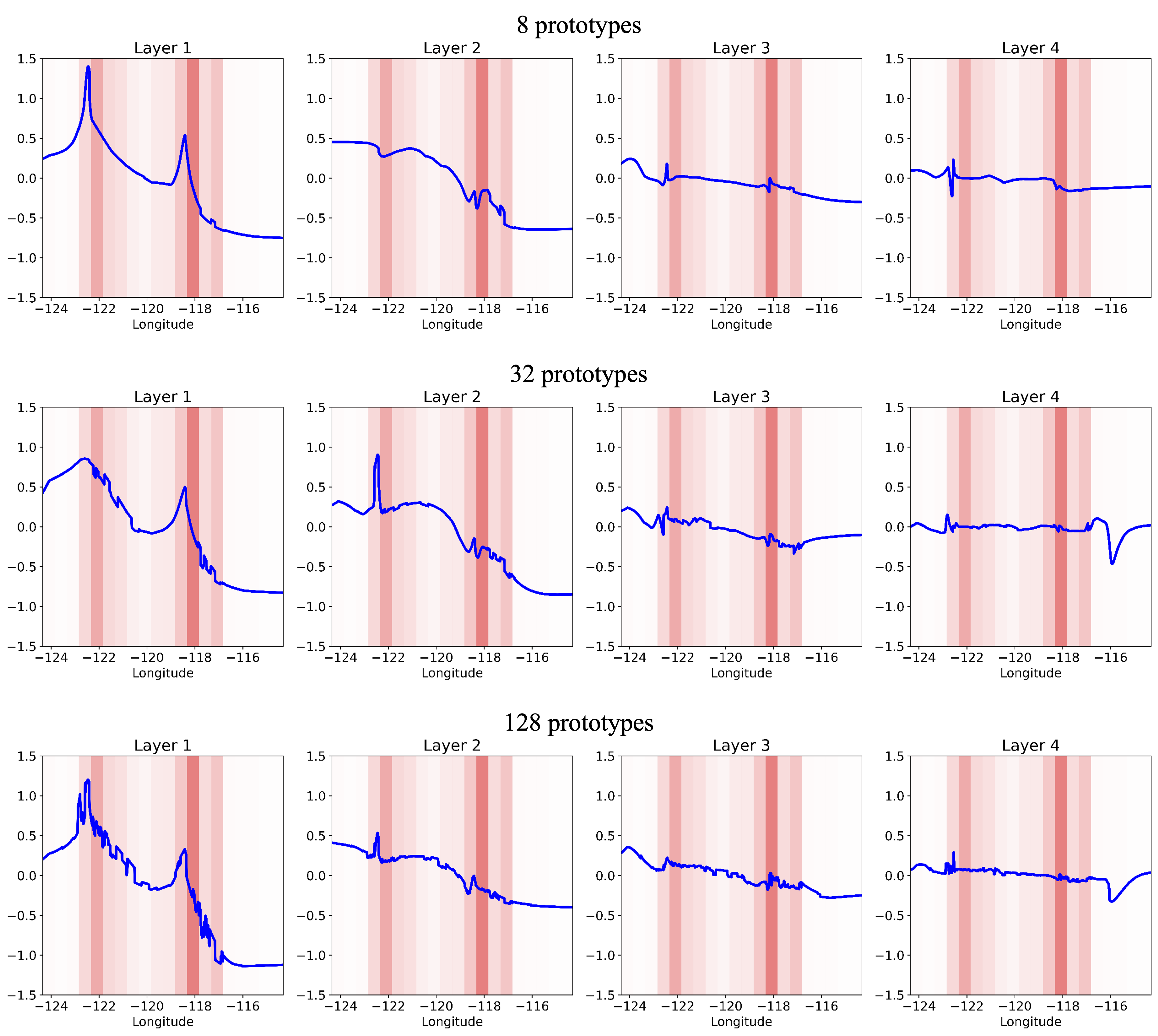}
    \caption{Shape plots of the ``Longitude" feature generated by ProtoNAM with different numbers of prototypes.}
    \label{fig:app_housing_n_proto}
\end{figure*}

As we introduced in Section \ref{sec:interpret}, ProtoNAM is able to draw the shape plots of each feature in tabular datasets, with $x$-axis as the feature value and $y$-axis as the mean-centered prediction output given by each feature.
Figure \ref{fig:app_housing_n_proto} shows the shape plots of ``Longitude'' feature generated by ProtoNAMs with different numbers of prototypes mentioned in Section \ref{sec:ablation}. It can be observed that as the number of prototypes increases, the model can learn more jagged shape functions, which results in better performance on the tasks as shown in Table \ref{tab:ablation_layer_proto}.

\subsection{Visualization of Binary Feature Interactions.}

\begin{figure*}[ht]
    \centering
    \includegraphics[width=0.8\linewidth]{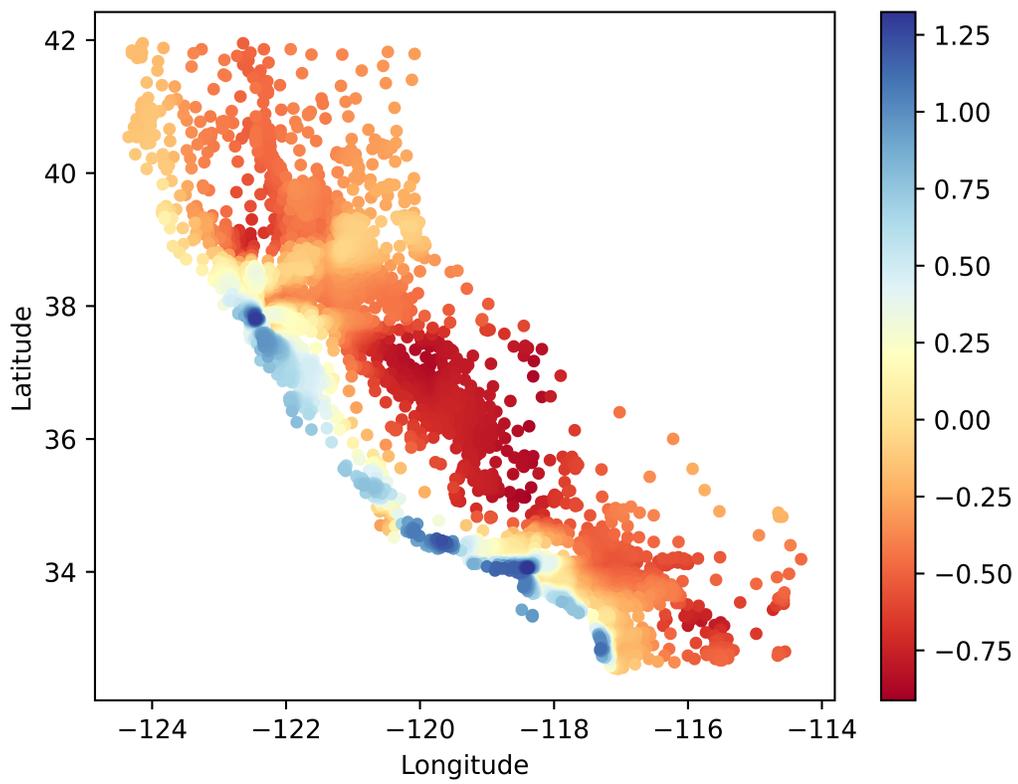}
    \caption{Interpretation of the Longitude-Latitude interaction in Housing provided by ProtoNA$^2$M.}
    \label{fig:app_bin_lat_lon}
\end{figure*}

\begin{figure*}
    \centering
    \includegraphics[width=\linewidth]{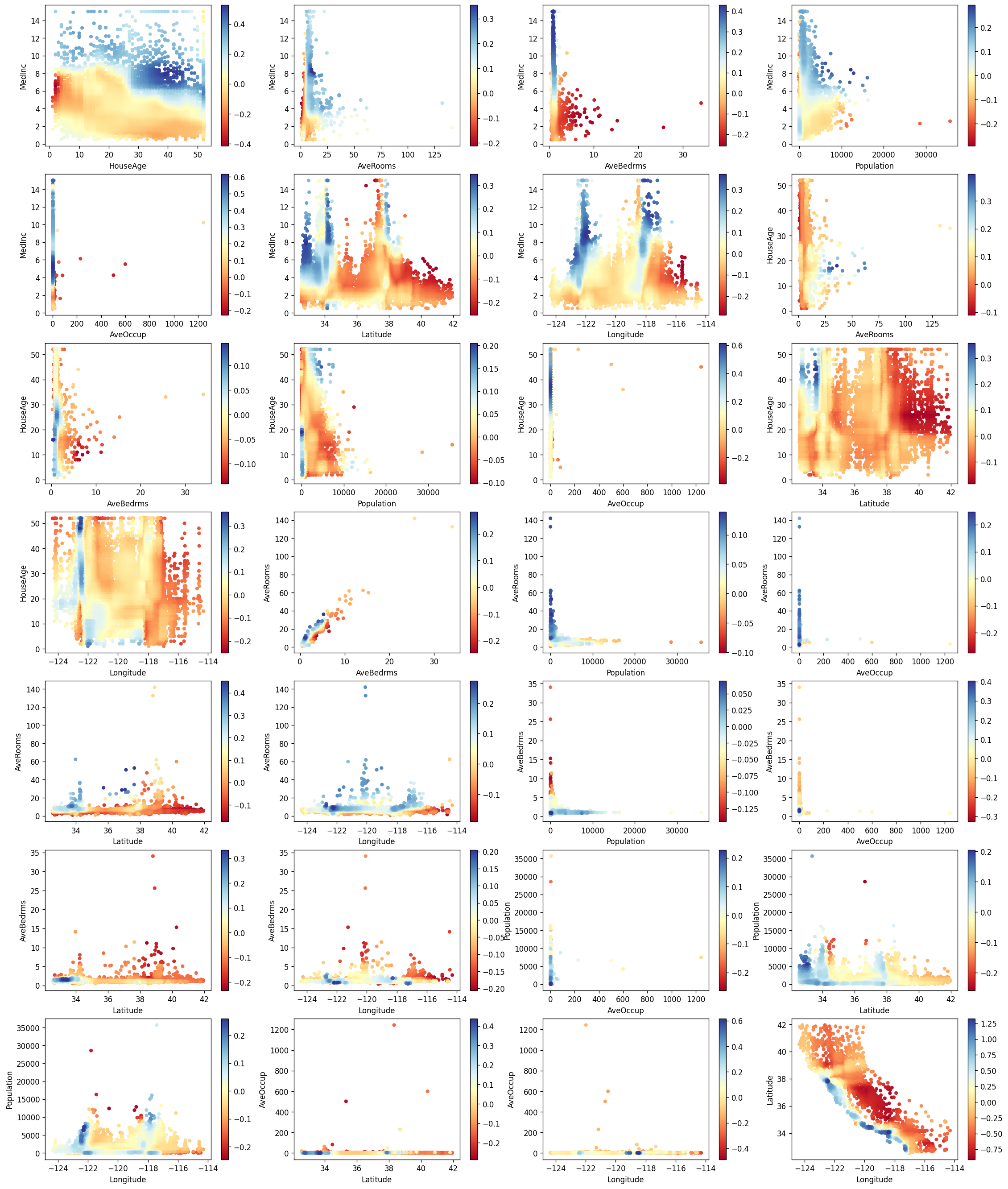}
    \caption{Interpretation of binary feature interactions in Housing provided by ProtoNA$^2$M.}
    \label{fig:app_bin_all}
\end{figure*}

{For ProtoNA$^2$M, we visualized the interpretation of feature interactions with scatterplots, where the predicted values are reflected by the colors. Figure \ref{fig:app_bin_lat_lon} shows the visualization of the interaction between Longitude and Latitude. We can observe that, in addition to San Francisco (37.8° N, 122.4° W) and Los Angeles (34.1° N, 118.2° W), the two areas where ProtoNAM predicts higher prices, the interpretation provided by ProtoNA$^2$M finds two other areas, Santa Barbara (34.4° N, 119.7° W) and San Diego (32.7° N, 117.2° W), where the model gives relatively high predicted values. More visualization results of the binary feature interactions of Housing dataset are shown in Figure \ref{fig:app_bin_all}.}

\subsection{Visualization of More Shape Functions.}

\begin{figure*}[h]
    \centering
    \includegraphics[width=0.8\linewidth]{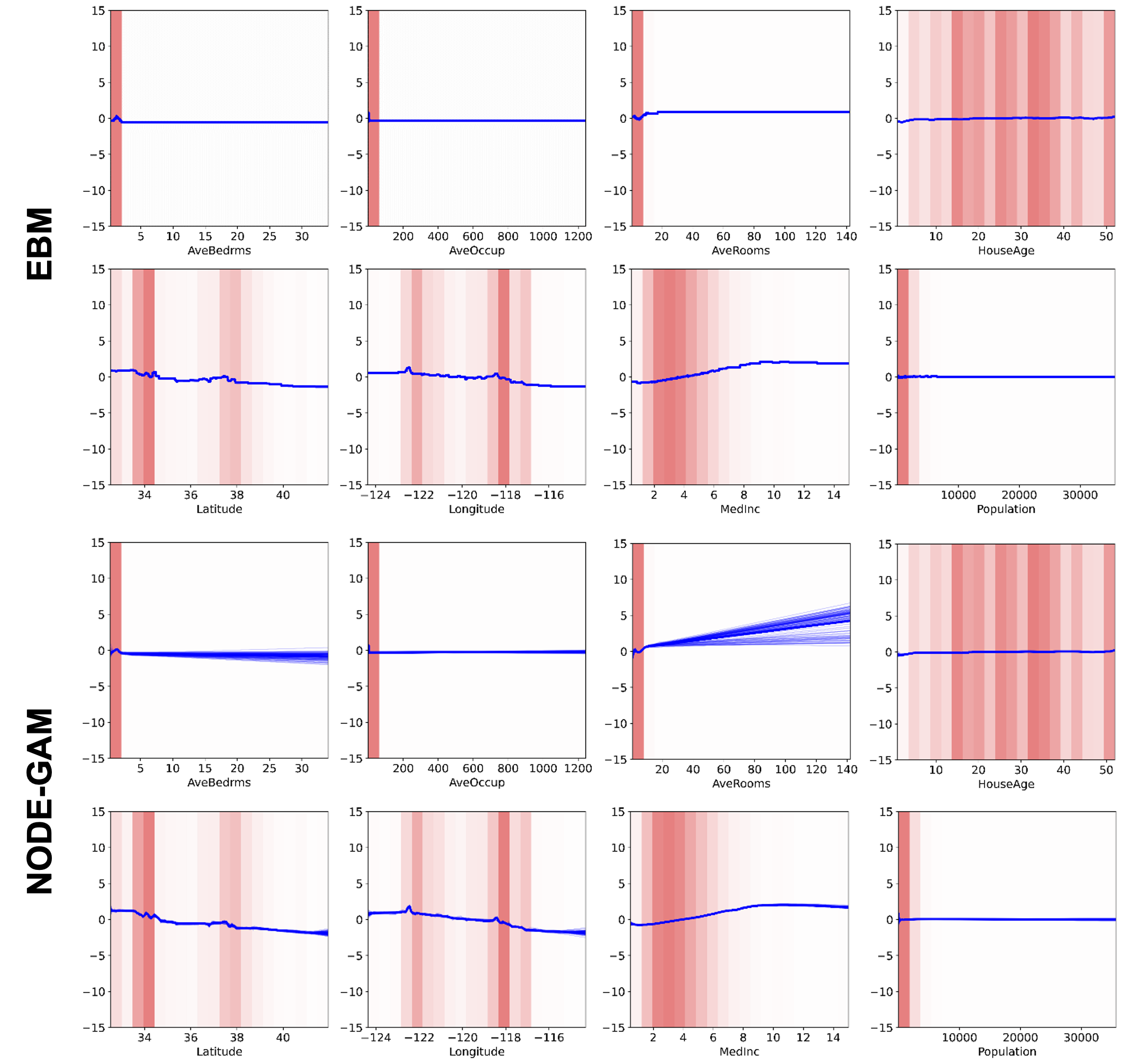}
    \caption{Shape plots of features in Housing generated by EBM and NODE-GAM.}
    \label{fig:app_housing_all_1}
\end{figure*}

\begin{figure*}[h]
    \centering
    \includegraphics[width=0.8\linewidth]{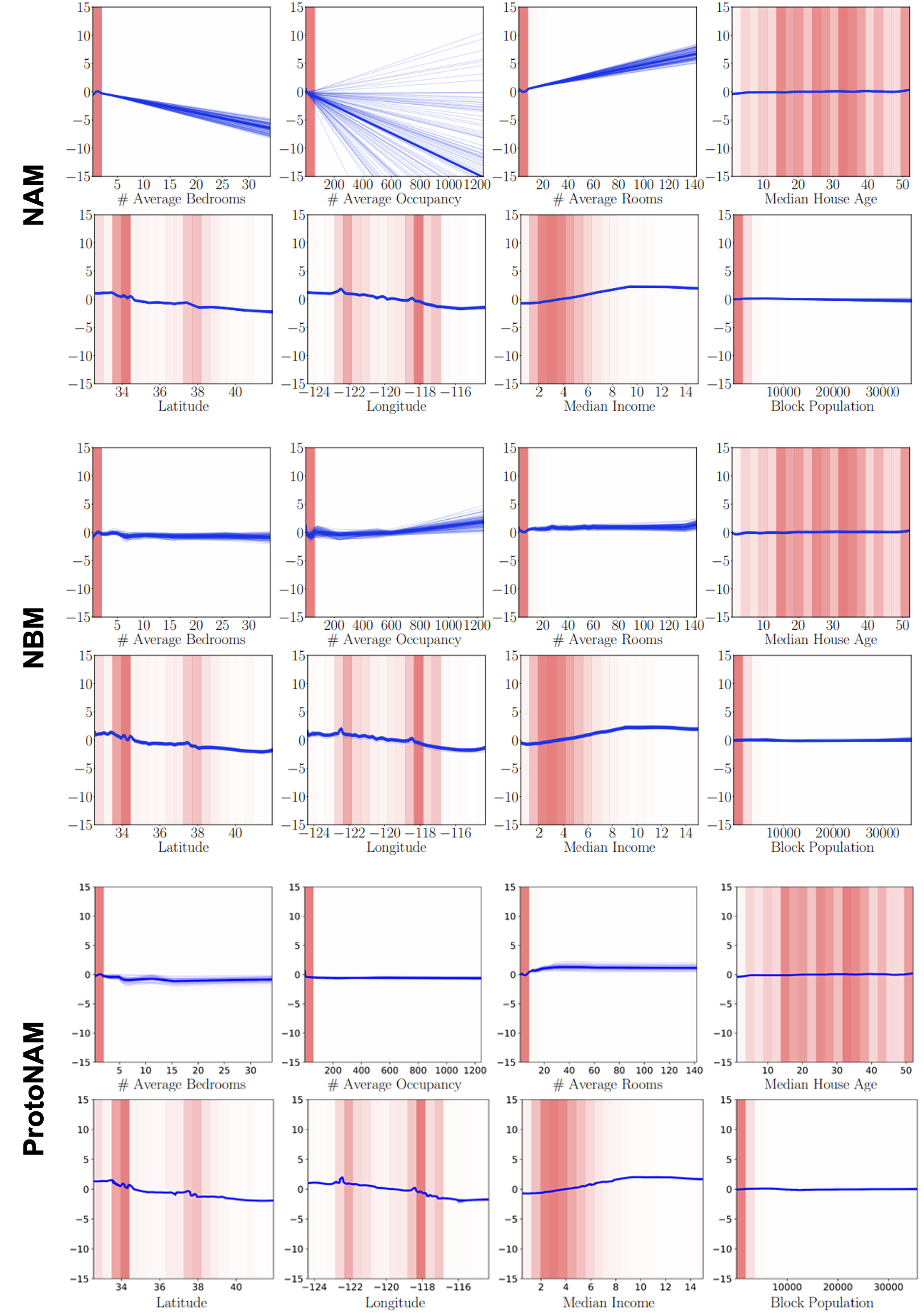}
    \caption{Shape plots of features in Housing generated by NAM, NBM, and ProtoNAM.}
    \label{fig:app_housing_all_2}
\end{figure*}

Since the number of features is only 8 in Housing, we plot shape functions of features in the dataset and compare them with all GAM baselines we used, as shown in Figures \ref{fig:app_housing_all_1} and \ref{fig:app_housing_all_2}. We can see from the figure that our model is both expressive for modeling irregular functions and robust for providing stable results. Moreover, from the plots of ProtoNAM we can observe that ``Latitude'', ``Longitude'', and ``Median Income'' are the three key features that affect the prediction of housing prices, suggesting that local housing prices are most closely related to the location and per capita income.

\begin{figure*}[h]
    \centering
    \includegraphics[width=0.8\linewidth]{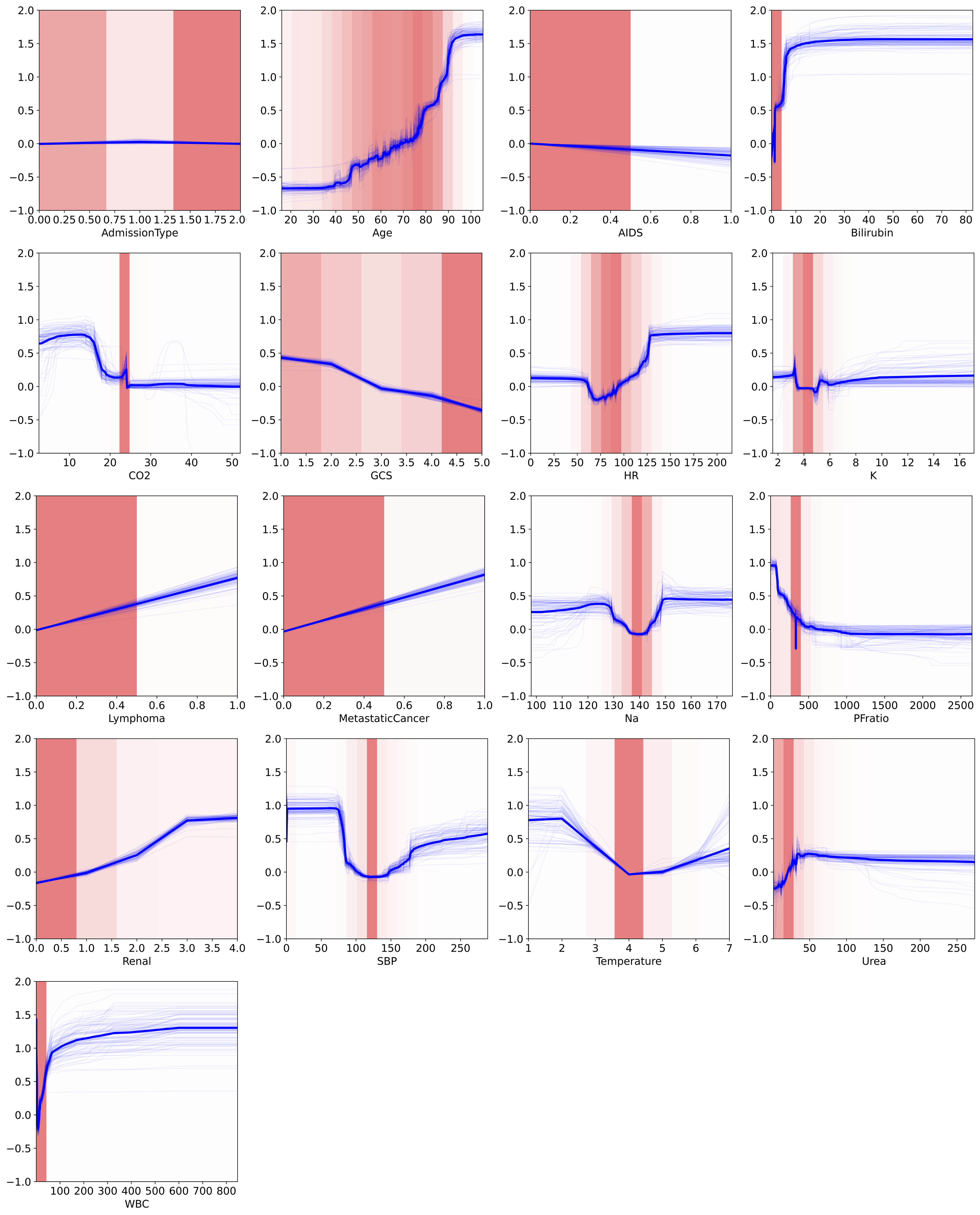}
    \caption{Shape plots of features in MIMIC-II generated by ProtoNAM}
    \label{fig:app_mimic2_shape_all}
\end{figure*}

\begin{figure*}[h]
    \centering
    \includegraphics[width=0.8\linewidth]{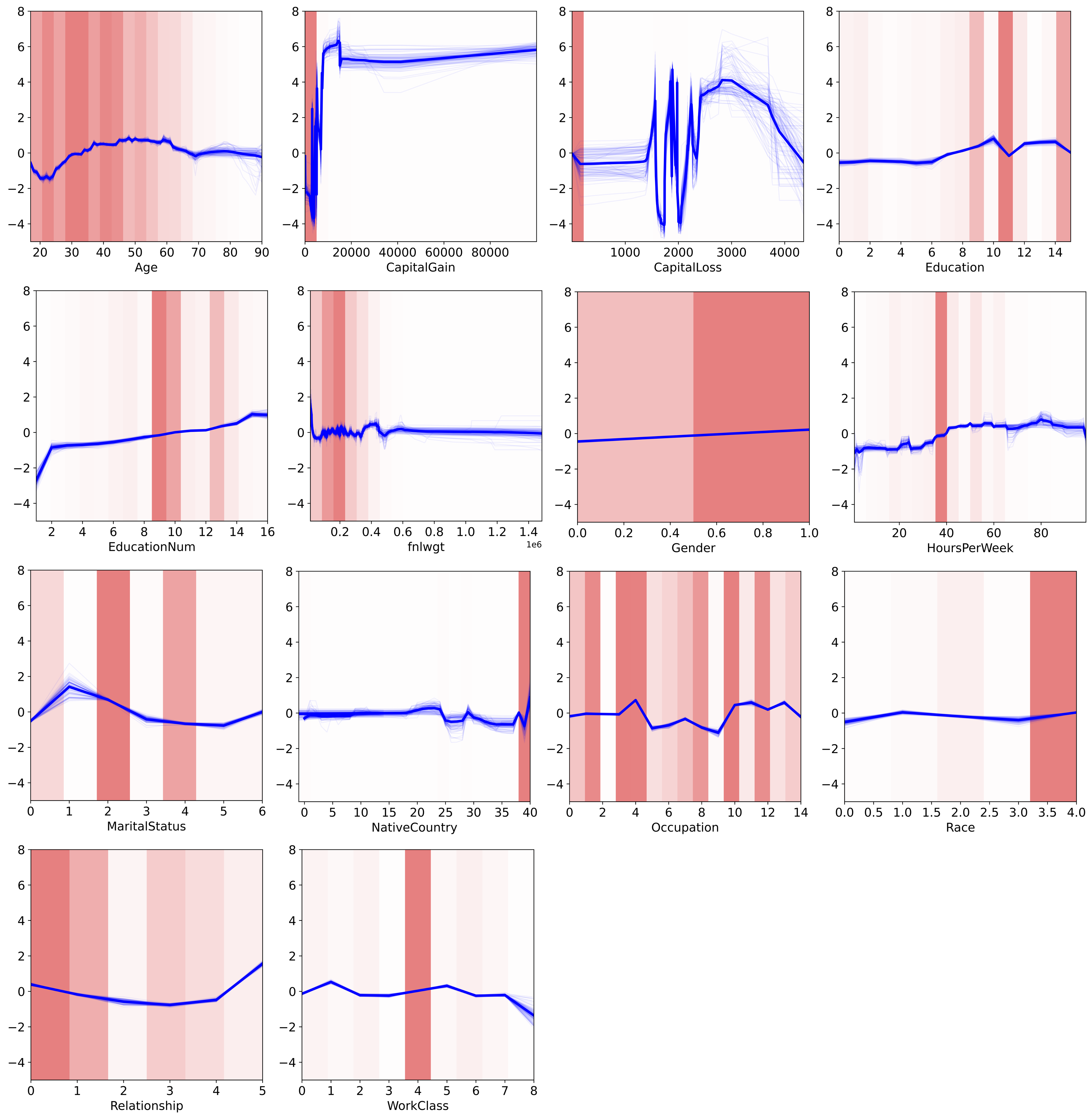}
    \caption{Shape plots of features in Income generated by ProtoNAM}
    \label{fig:app_income_shape_all}
\end{figure*}

We provide additional visualization results for shape plots of features in MIMIC-II (Figure \ref{fig:app_mimic2_shape_all}) and Income (Figure \ref{fig:app_income_shape_all}) datasets. Form Figure \ref{fig:app_mimic2_shape_all}, we can see that ProtoNAM identifies the feature ``Age'' as the feature with the greatest variability in impact on prediction and also detects many small jumps in its shape function. Such sharp variations are also detected in the shape functions of many other features such as ``Billirubin'', ``CO$_2$'', ``K'', ``PFratio'', which provide us with a precise insight into the link between the mortality and these features.

Figure \ref{fig:app_income_shape_all} shows the complex patterns discovered by ProtoNAM which are embedded in the features of the Income dataset. As can be seen, the effect of the feature ``Age'' on Income prediction is highest between the ages of 50 to 60 and decreases thereafter. It is also interesting to see that a sample with a larger ``EducationNum'' will have a higher probability to earn over \$50K per year.
In addition, we observe that the values of the shape functions vary considerably in the plots of ``CaptialGain'' and ``CapitalLoss'', indicating that the relationship between the prediction and these features is quite complex that can differ significantly across samples.

\section{Limitations and Future Work}

From Figures \ref{fig:app_housing_all_2}, \ref{fig:app_mimic2_shape_all}, \ref{fig:app_income_shape_all}, we see that though ProtoNAM can learn irregular data patterns, it may fail to do so if samples are few in local regions, as fewer prototypes would be learned in regions with a low data density. This is a general problem of data imbalance that can be encountered by any model, which is beyond the scope of this study. In addition, involving human experts in the evaluation of the explanations could be a valuable future direction, which facilitates a comparative analysis of the explanations provided by the different GAMs.